\documentclass[10pt,journal,compsoc]{IEEEtran}
\ifCLASSOPTIONcompsoc
  \usepackage[nocompress]{cite}
\else
  \usepackage{cite}
\fi

\ifCLASSINFOpdf
\else
\fi

\usepackage{graphicx}
\usepackage{amsmath}
\usepackage{amsfonts}
\usepackage{bm}
\usepackage{makecell}
\usepackage{indentfirst,booktabs,multirow,tabularx}
\usepackage[table]{xcolor}
\usepackage[noend]{algpseudocode}
\usepackage{algorithmicx,algorithm}
\usepackage[colorlinks,bookmarksopen,bookmarksnumbered,citecolor=green, linkcolor=blue, urlcolor=red]{hyperref}
\usepackage{subfigure}

\usepackage{balance}

\definecolor{OverviewDatabase}{RGB}{0,176,80}
\definecolor{OverviewMethod}{RGB}{0,112,192}
\definecolor{Checked}{RGB}{0,176,240}

\definecolor{TableLine}{RGB}{234,234,254}
\begin{document}
%
\title{Blind Image Quality Assessment: A Brief Survey}
\author{Miaohui~Wang,~\IEEEmembership{Senior Member,~IEEE}

\IEEEcompsocitemizethanks{\IEEEcompsocthanksitem M. Wang is with the State Key Laboratory of Radio Frequency Heterogeneous Integration and the Guangdong Key Laboratory of Intelligent Information Processing, Shenzhen University, Shenzhen 518060, China.\protect\\E-mail: wang.miaohui@gmail.com
}
}

%


\IEEEtitleabstractindextext{%
\begin{abstract}
Blind Image Quality Assessment (BIQA) is essential for automatically evaluating the perceptual quality of visual signals without access to the references. In this survey, we provide a comprehensive analysis and discussion of recent developments in the field of BIQA. We have covered various aspects, including hand-crafted BIQAs that focus on distortion-specific and general-purpose methods, as well as deep-learned BIQAs that employ supervised and unsupervised learning techniques. Additionally, we  have explored multimodal quality assessment methods that consider interactions between visual and audio modalities, as well as visual and text modalities. Finally, we have offered insights into representative BIQA databases, including both synthetic and authentic distortions. We believe this survey provides valuable understandings into the latest developments and emerging trends for the visual quality community.
\end{abstract}

}

\maketitle

\IEEEdisplaynontitleabstractindextext

%
\IEEEpeerreviewmaketitle

\IEEEraisesectionheading{\section{Introduction}\label{sec:introduction}}

\begin{figure*}[!t]
\centering
\includegraphics[width=0.85\textwidth]{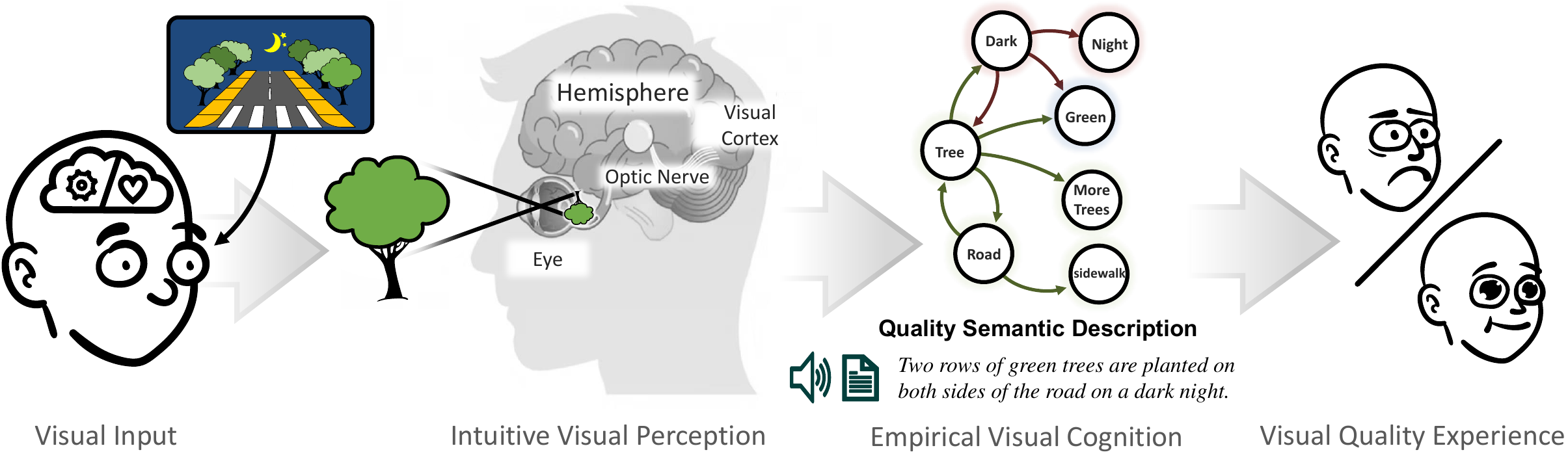}\\
\caption{\textit{Illustration of multimodal quality assessment}. Humans perceive the world through five sensory mechanisms, which are sight, hearing, touch, taste, and smell \cite{baltruvsaitis2018multimodal}. Stimulus information from different senses interacts to generate human perception and cognition \cite{pinson2012influence,min2020multimodal}. Consequently, both visual and text or auditory modalities can be considered in the visual quality assessment task.}
\label{fig:facade}
\end{figure*}

\IEEEPARstart{B}{lind} image quality assessment (BIQA) aims to automatically and accurately estimate objective visual scores, thereby overcoming the challenges of subjective experiments, such as being time-consuming, unstable, and non-automated \cite{wang2022low}. It plays a significant role in monitoring the quality of industrial products \cite{teixeira2016new}. In the BIQA task, the human visual system (HVS) is the final receiver of visual signals \cite{teixeira2016new,wang2021active,zhang2022continual}, but human visual perception is essentially the joint action of multiple sensory information \cite{min2020study}. However, existing hand-crafted and deep-learned BIQA methods rarely consider multimodal information \cite{cao2023subjective}, and their ability to measure complex image quality is limited. As a result, exploring how to utilize multimodal learning to enhance the accuracy of visual quality assessment is a promising research direction, as illustrated in Fig. \ref{fig:facade}.

When scoring the quality of visual signals, humans can perceive multiple sensing information (\textit{e.g.}, sight, hearing, \textit{etc}) simultaneously \cite{pinson2012influence,baltruvsaitis2018multimodal,cao2023subjective}. 
After acquired exercise, our brains can easily make connections between different modality data and further construct a comprehensive representation of visual characteristics \cite{song2019harmonized}. For example, when the image modality is influenced by various artifacts, other auxiliary modalities, such as audio-based ambient information \cite{cao2023subjective} and text-based visual understanding \cite{wang2023blind}, are expected to provide additional clues for describing quality. Consequently, BIQA aims to establish a visual indicator that mimics the manner of human visual perception, and it is expected to learn better quality descriptors to represent visual perception.

\subsection{Related BIQA Surveys}
Objective image quality assessment (IQA) \cite{sun2023blind} has been widely applied to visual communication systems. This is due to the degradation introduced in visual signal acquisition, compression, processing, transmission, and display. Objective IQA is favored for its efficiency, and numerous studies have explored subjective and objective visual quality assessment.  Traditionally, IQA can be categorized into three types: full-reference, reduced-reference, and no-reference. No-reference IQA, also known as BIQA, specifically refers to the evaluation of image quality without access to the original reference, which is a challenging and important task in real-world scenarios \cite{hou2023uid2021}. It is worth noting that Zhai \textit{et al.} \cite{zhai2020perceptual} reviewed recent developments on full-reference IQA methods, while Dost \textit{et al.} \cite{dost2022reduced} provided the latest survey on reduced-reference IQA methods. In this section, we will mainly review the surveys for BIQAs as follows.

In 2015, Manap \textit{et al.} \cite{manap2015non} reviewed BIQA from the perspective of non-distortion-specific cases. This  survey categorized existing approaches into natural scene statistics-based and learning-based BIQAs. Additionally, they discussed the performance, limitations, and research trends.

In 2017, Xu \textit{et al.} \cite{xu2017no} presented a brief review of BIQA, with a focus on general-purpose algorithms. They  covered the fundamental developments in BIQA, categorizing them into distortion-specific and general-purpose categories. They also discussed recent progress in feature extraction and quality prediction. Moreover, they have conducted comparisons on several benchmark databases and highlighted the ongoing BIQA challenges.

In 2019, Yang \textit{et al.} \cite{yang2019survey} provided a survey of BIQA methods, focusing on recent developments in deep neural networks (DNNs). Performance comparisons on both synthetic and authentic databases (\textit{e.g.}, LIVE, TID2013, CSIQ, LIVE multiply distorted, LIVE challenge) have offered valuable insights into the strengths and limitations of various DNN-based BIQAs.  This survey summarized the evolving landscape of DNN-based BIQA methods, demonstrating the intrinsic relationships among different BIQAs, and discussed the challenges and directions for future research.


In 2022, Stepien \textit{et al.} \cite{stkepien2022brief} reviewed recent developments of BIQA methods specific to magnetic resonance imaging (MRI). They discussed the common distortions found in MR images, categorized popular BIQA methods, and outlined existing approaches in describing MRI images and developing quality prediction models. Evaluation protocols and benchmark databases were also discussed. Despite the limited number of studies focusing on MR image assessment, this work   highlighted diverse approaches, with a particular focus on the increasing importance and main challenges of deep learning architectures.

\subsection{Necessity of New BIQA Surveys}
BIQA has attracted significant attention due to the increasing demand for high-quality user experience \cite{amini2023towards} across various domains, including broadcasting, remote education, game streaming, social media, and more. However, existing BIQA surveys face challenges in thoroughly exploring image quality models. It suggests the urgency for a comprehensive survey that can effectively identify emerging trends and challenges in the rapid evolution of BIQA technologies.

Firstly, one of the main reasons for the new BIQA survey is to adapt to rapidly developing imaging technologies. With the advancement in camera sensors and display technologies, the content of digital images are becoming more diverse and complex (\textit{e.g.}, screen content, user-generated content, 360$^\circ$ visual content, \textit{etc}). Therefore, traditional BIQA models that rely on simplistic features or assumptions may struggle to capture the characteristics of image quality, leading to inaccurate assessments. A comprehensive survey can help to reveal the associated challenges for BIQAs and the emerging trends in imaging technologies.

Secondly, diverse applications have also increased the demand for quality assessment of application-specific content, making BIQA important to measure the perceptual quality of distortion-specific images without explicit knowledge of the reference images. However, the performance of distortion-specific BIQA models under real-world scenarios, such as low-light conditions, immersive display devices, and visual-audio preferences, remains a concern. Conducting a new BIQA survey can update the overview of state-of-the-art  methods and their ability to deliver accurate quality assessments.

Thirdly, image processing has undergone significant changes due to the proliferation of deep learning tools, which have demonstrated remarkable capabilities in learning intricate quality features. By leveraging large-scale datasets and complex network architectures, deep learned BIQA models can effectively capture subtle visual cues and patterns that are indicative of image quality. Moreover, deep learning paradigms, such as meta learning, transfer learning, and multimodal learning, have been employed to represent different types of distortions and diverse image contents in BIQA. Despite these advancements, BIQA still faces learning-based challenges, including domain adaptation, robustness to various distortions, and generalization to diverse image datasets. Therefore, ta new survey summarizing the latest deep learning methods, including meta-learning, transfer learning, and CNN-based approaches, in the context of no-reference scenarios is highly warranted. Such a survey can provide valuable insights into the strengths and limitations of existing deep learning-based BIQA models.

To sum up, there is a clear need for a new BIQA survey due to the rapid evolution of imaging technologies, the widespread use of visual applications, and the increasing availability of intelligent tools. Conducting a comprehensive survey that reviews the latest research findings, identifies emerging challenges, and inspires future directions in BIQA methods will be beneficial to researchers, practitioners, and industry professionals in the field of visual quality assessment.

\subsection{Scope of This Survey}
In this survey, we focus on recent developments of BIQA, and provide a brief analysis and discussion in various aspects, covering hand-crafted BIQA methods (\textit{e.g.}, distortion-specific and general-purpose), deep-learned BIQA methods (\textit{e.g.}, supervised learning-based and unsupervised learning-based), multimodal quality assessment methods (\textit{e.g.}, visual-audio and visual-text), and representative databases (\textit{e.g.}, synthetic and authentic distortion).

\section{Hand-crafted BIQA Methods}\label{subsec:traditional_method}
Hand-crafted BIQA methods \cite{cai2023blind} typically rely on the feature extraction techniques derived from expert and engineering experience. These methods are designed to capture the characteristics or attributes of image quality based on domain knowledge.  Hand-crafted BIQAs often require fewer resources in terms of database size, hardware platform, and computing power, and they  highly achievable and easy to deploy in practical applications. Based on the specific application context, existing BIQA methods can be categorized into two main types: 1) distortion-specified and 2) general-purpose methods. 
\begin{figure*}[!t]
\centering
\includegraphics[width=0.99\textwidth]{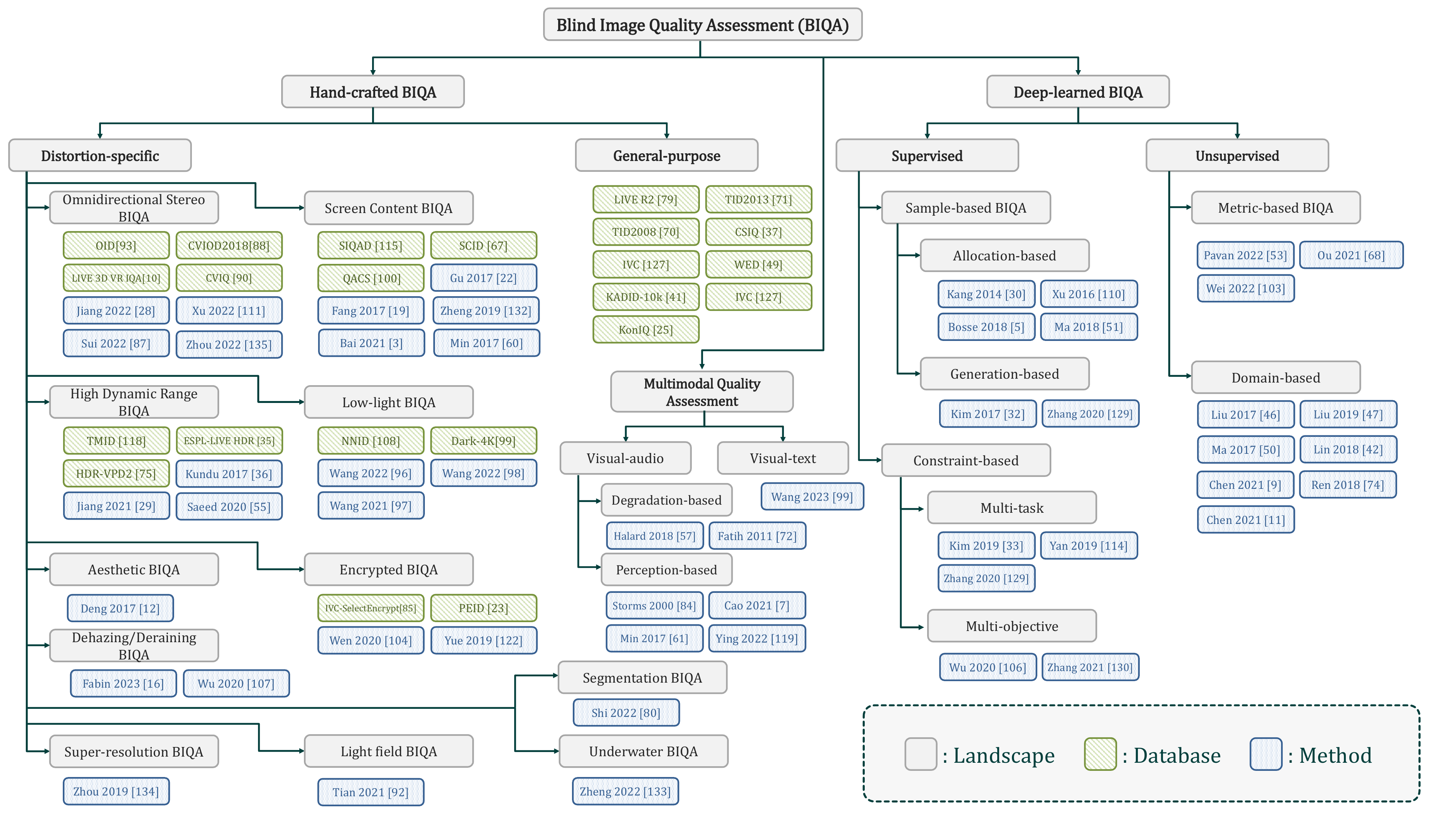}\\
\caption{\textit{A concise illustration of blind image quality assessment (BIQA)}. Representative methodologies are briefly demonstrated from the perspective of hand-crafted BIQA methods (\textit{e.g.}, distortion-specific and general-purpose) and deep-learned BIQA methods (\textit{e.g.}, supervised learning-based and unsupervised learning-based).}
\label{fig:overview_related_work}
\end{figure*}

\subsection{Distortion-specific BIQAs}\label{subsec:distortion_specified_method}
Distortion-specific BIQAs \cite{yu2023hybrid} quantify the image quality by considering both degradation manners and distortion types of a particular application, as shown in Fig. \ref{fig:overview_related_work}. In this section, we present a concise overview of several representative applications and the significant milestones in the field.

\subsubsection{\textbf{Screen Content}}
The development of computer-generated technology has promoted the widespread of screen content (SC) visual signals \cite{min2021screen}, and further drives the requirement of SC-based BIQA. Various SC-based databases have been established to facilitate the investigation in this area, including Screen Image Quality Assessment Database (SIQAD) \cite{yang2015perceptual}, Screen Content Database (SCD)\cite{shi2015study}, quality assessment of compressed SCI database (QACS) \cite{wang2016subjective}, and Screen Content Image Database (SCID) \cite{ni2017esim}.  These databases provide valuable resources for evaluating the quality of SC images. Given the distinct features of SC images, many representative BIQAs have been proposed, such as structural feature-induced \cite{gu2017no,min2017unified},  brightness and texture-driven \cite{fang2017no,fang2020perceptual}, region division-based \cite{zheng2019no,bai2021blind}, \textit{etc}. The continued development of SC-based BIQA methods is crucial for ensuring the quality of SC-based visual signals, which is prevalent in various domains such as multimedia streaming, video conferencing, and remote education.

\subsubsection{\textbf{Low-light}}
Imaging in weak-illumination environments can lead to uneven brightness, poor visibility, impaired color, and increased hybrid noise. These factors not only degrade the user experience but also affect the product value \cite{wang2022low}. Recognizing the significance of addressing low-light distortion, researchers have developed specialized low-light databases and BIQA methods. The Natural Night-time Image Database (NNID) \cite{xiang2020blind} is one of the earliest low-light BIQA databases, and later a new ultra-high-definition low-light database, Dark-4K, has been established  \cite{wang2023blind}. To tackle the challenges associated with modeling low-light distortions, some preliminary BIQAs have been developed for low-light images, such as brightness-guided \cite{xiang2020blind}, colorfulness-inspired \cite{wang2021blind}, visibility-induced \cite{wang2022low}, and comparative learning-based \cite{wang2022super}. These efforts in developing low-light BIQAs is crucial in improving the assessment of image quality in weak-illumination scenarios, which can unlock the full potential of imaging technologies in challenging lighting conditions.

\subsubsection{\textbf{High Dynamic Range (HDR)}}
HDR-based BIQAs are designed to evaluate the quality degradation primarily caused by tone mapping or multiple exposure fusion techniques in High Dynamic Range (HDR) images. To facilitate research in this region, several representative HDR databases have been established, such as Tone-mapped image database (TMID) \cite{yeganeh2012objective}, HDR JPEG XT database (HDR-XT) \cite{korshunov2015subjective},  HDR compression database (HCD) \cite{zerman2017extensive},  ESPL-LIVE HDR image database \cite{kundu2017large},  and HDR visual difference predictor 2 (HDR-VPD2) \cite{rousselot2018impacts}. 
In the development of HDR-based BIQAs, various quality assessment methods have been proposed, taking into consideration representative quality descriptors such as gradient \cite{kundu2017no}, color \cite{saeed2020multi}, and brightness \cite{jiang2021blind}. These methods aim to capture the visual distortions introduced during the tone mapping or multiple exposure fusion processes and quantify their impact on the perceived quality of HDR images.
With the continuous development of HDR, the use of new technologies (tone-mapping or exposure fusion algorithm) has improved HDR imaging results. It remains to be verified whether existing methods can accurately measure the distortion introduced by new HDR technologies.

\subsubsection{\textbf{Encryption}}
In secure communication, copyright protection, and privacy preservation, the need to guarantee the quality experience for encrypted images has driven the development of encryption-based BIQA. These methods focus on evaluating the quality of images that have been encrypted to protect them. In this context, two representative databases have been established: the IVC-SelectEncrypt database\cite{stutz2010subjective} and the perceptually encrypted image database (PEID)\cite{guo2019peid}. 
Different from traditional BIQAs, the quality metrics for encrypted images require a unique consideration: striking a balance between maintaining a relatively high level of quality to attract unauthorized users while ensuring security from unauthorized viewers \cite{yue2019no,wen2020visual}.  Encryption-based BIQAs aim to address this dual objective by developing quality metrics for encrypted images.

\subsubsection{\textbf{Omnidirectional Stereo}}
The popularity of immersive multimedia applications has promoted the development of stereoscopic-omnidirectional BIQA tasks. Several benchmark databases have been established to address this need, including the omnidirectional image database (OID) \cite{upenik2016testbed}, Compressed 360-degree Image Database (CVIQD2018) \cite{sun2018large}, Omnidirectional IQA (OIQA) \cite{duan2018perceptual},  Head-mounted Immersive Image Database (HIID)  \cite{huang2018modeling}, IISc Stitched Image QA (ISIQA)  \cite{madhusudana2019subjective}, LIVE 3D VR IQA \cite{chen2020study}, Compressed VR Image Quality (CVIQ) \cite{sun2020mc360iqa}, Multi-Distortions Visual Attention Quality Database (MVAQD) \cite{jiang2021cubemap}, Omnidirectional Image Quality (OIQ)  \cite{fang2022perceptual}, and the psychophysical-related database  (PRD) \cite{sui2022perceptual}. 
One of the main paradigms for quality feature extraction is based on multi-view  decomposition \cite{huang2018modeling,jiang2021cubemap,zhou2022no}.
However, in addition to the quality of the visual signal itself, there are many factors that affect the  perceived quality of virtual reality experiences, such as viewing  behavior \cite{sui2022perceptual} and camera motion \cite{xu2022viewport}. Therefore, there is still significant room for exploring and developing new quality descriptors specifically tailored to the stereoscopic-omnidirectional scenario.

\subsubsection{\textbf{Other Applications}}
Besides the application mentioned above, there are several other promising distortion-specific BIQAs  that have been designed for emerging computer vision tasks, such as super-resolution \cite{zhou2019visual}, segmentation \cite{shi2022objective}, dehazing \cite{erlenbusch2023thermal}, deraining \cite{wu2020subjective}, aesthetic \cite{deng2017image}, light-field \cite{tian2020light}, underwater \cite{zheng2022uif}, 3D geometry \cite{wang2022combining}, \textit{etc}. Note that each of these tasks presents unique challenges and requires specialized BIQA methods designed to the specific distortion type. Due to the length of the article, we will not discuss other types of distortion-specific BIQAs here. A brief illustration of the related work is organized as shown in Fig. \ref{fig:overview_related_work}.

\subsection{General-purpose BIQAs}
Unlike distortion-specific methods tailored to address particular types of distortion, general-purpose BIQAs \cite{zhao2023quality} are not designed with the typical image applications mentioned above. These methods usually utilize employ a set of common quality-aware features to assess image distortion, which can be roughly divided into two main groups: natural scene statistics (NSS) based methods and HVS-guided methods. In this subsection, we provide a concise overview of some representative general-purpose BIQAs and databases.

\subsubsection{\textbf{NSS-based}}  
NSS-based BIQAs are based on the assumption that high-fidelity images obey specific statistical characteristics that are altered by quality degradation \cite{moorthy2011blind}. For instance, early methods quantified these alterations in the frequency domain, such as  wavelet transform \cite{moorthy2011blind} or discrete cosine transform  \cite{saad2012blind}. However, due to the computational complexity associated with domain transformations, some methods instead directly select descriptive quality-aware features in the spatial domain,  such as normalized luminance \cite{mittal2012no}, gradient magnitude \cite{xue2014blind}, and local binary pattern \cite{liu2020blind}. To better capture the NSS characteristics, parametric models have also been developed,  such as generalized Gaussian density (GGD) \cite{moorthy2010two}, multivariate GGD \cite{mittal2013making}, and asymmetric GGD \cite{zhang2015feature}. It is worth noting that existing NSS-based BIQAs usually struggle to fully consider the distortion characteristics of different types of image distortions, so it is difficult for such methods to achieve the optimal performance in practical applications.

\subsubsection{\textbf{HVS-guided}} 
Considering that humans are the ultimate recipients of visual signals, it is significant to leverage the perceptual characteristics of the HVS in the design of image quality indicators.  HVS-guided BIQAs  aim to represent these characteristics and enhance the assessment of image quality. Two prominent categories of HVS-guided methods are the free-energy principle-driven and the visual sensitivity-based methods. The free-energy principle interprets the quality perception in visual signals as an active inference process \cite{zhai2012psychovisual}, which provides a theoretical framework for understanding the perception of image quality. On the other hand, the visual characteristics-based BIQAs primarily rely on visual sensitivity features  to assess image quality, such as luminance \cite{li2016blind}, structure \cite{li2017bsd}, texture \cite{liu2018no}, and  color \cite{freitas2018no}. By considering the specific HVS sensitivities, these methods provide a more accurate evaluation of image quality. However, one of the main challenges in HVS-guided BIQAs is the complexity of general-purpose distortions, which can be affected by various factors. Simple HVS models and visual sensitivities may struggle to adequately characterize and capture the intricacies of these distortions. Thus, further research is needed to develop more sophisticated models and visual features that can effectively address the complex distortions in HVS-guided BIQAs.
\begin{table*}[!t]
\centering
\caption{\textbf{Summary of representative databases used in the BIQA task}.}
\label{tab:database_summary}
\renewcommand{\arraystretch}{1.5}
\setlength{\tabcolsep}{8.4mm}{
\begin{tabular}{c|c|c}
\toprule[1.5pt]
Category &Synthetic database &Authentic database\\
\hline
\makecell{Distortion-\\specific} &\makecell{SIQAD \cite{yang2015perceptual}, ~SCD \cite{shi2015study}, ~QACS \cite{wang2016subjective}, ~SCID \cite{ni2017esim},
\\{TMID \cite{yeganeh2012objective}, ~HDR-XT \cite{korshunov2015subjective}, ~HCD \cite{zerman2017extensive},~~}
\\ESPL-LIVE~HDR \cite{kundu2017large},~HDR-VPD2 \cite{rousselot2018impacts},
\\OID \cite{upenik2016testbed},~CVIQD2018 \cite{sun2018large}, ~OIQA \cite{duan2018perceptual}, ~HIID \cite{huang2018modeling},
\\{ISIQA \cite{madhusudana2019subjective}, ~LIVE~3D~VR~IQA \cite{chen2020study},~CVIQ \cite{sun2020mc360iqa},~~~~}
\\MVAQD \cite{jiang2021cubemap}, ~OIQ \cite{fang2022perceptual}, ~PRD \cite{sui2022perceptual}, ~A/V-QA \cite{min2020study}}
&\makecell{NNID \cite{xiang2020blind},
\\{~~~~~~~\textbf{MLIQ} \cite{wang2023blind},~~~~~~~~~}
\\{~~~~~~~\textbf{Dark-4K} \cite{wang2023blind}~~~~~~~}}\\	
\hline		
\makecell{General-\\purpose}&\makecell{{LIVE \cite{sheikh2006image},~TID2008/2013  \cite{ponomarenko2009tid2008,ponomarenko2015image},~~~~}
\\{CSIQ \cite{larson2009categorical}, ~KADID-10k \cite{lin2019kadid}, ~WED \cite{ma2016waterloo}},
\\{IRCCyN/IVC~Database \cite{zhang2015application}~~~~~~~~~~~~~~}}
&KonIQ \cite{hosu2020koniq} \\
\bottomrule[1.5pt]
\end{tabular}}
\end{table*}

\subsubsection{\textbf{Databases}}
The performance evaluation of general-purpose BIQA methods depends heavily on the types of distortions  contained in IQA databases. Several commonly used databases for general-purpose BIQAs include the  Laboratory for Image $\&$ Video Engineering (LIVE) database \cite{sheikh2006image}, Tampere Image Database (TID) 2008/2013 \cite{ponomarenko2009tid2008,ponomarenko2015image}, Categorical Subjective Image Quality (CSIQ) database \cite{larson2009categorical}, Subjective Quality Assessment IRCCyN/IVC Database \cite{zhang2015application}, Waterloo Exploration Database (WED) \cite {ma2016waterloo}, Konstanz Artificially Distorted Image quality Database (KADID-10k) \cite{lin2019kadid}, KonIQ-10k \cite{hosu2020koniq}, and others.  These IQA databases consider various types of distortions, with the most prevalent ones being compression distortion  (\textit{e.g.}, JPEG, JPEG2K, H.26X), blur distortion (\textit{e.g.}, Gaussian blur, motion blur, out-of-focus blur), and different types of noises (\textit{e.g.}, white noise, impulse noise,  multiply noise).  Table \ref{tab:database_summary} provides a brief summary of some representative databases in terms of the distortion type (\textit{e.g.}, authentic and synthetic) and the application scenario (\textit{e.g.}, distortion-specific and general-purpose).

Existing general-purpose BIQAs generally achieve satisfactory performance on synthetic databases, but  their performance is relatively limited when faced with images captured in weak-illumination scenarios. As industrial products and services continue to evolve, the imaging quality of vision sensors in specific scenes (\textit{e.g.}, taking pictures at night) is more likely to be used as a product value indicator \cite{wang2022low}. Therefore, there is a practical significance in both industry and academia for the development of distortion-specific BIQA databases that feature authentic distortions, which can better reflect real-world scenarios and enable more accurate assessment of image quality in specific application domains.

\subsection{Summary}
While existing distortion-specific and general-purpose BIQAs have demonstrated reliable performance in their respective applications, it is highly desired to develop specialized quality indicators for authentic image distortions. This necessity arises for the following reasons.

Firstly, it is important to highlight the fundamental difference between synthetic and authentic distortions. Synthetic distortions are typically straightforward and can be easily controlled and manipulated during the generation of IQA datasets. On the other hand, authentic distortions are far more complex and can arise during various stages of image acquisition, compression, and processing in real-world scenarios.  Therefore, developing quality indicators specific to authentic distortions holds greater practical value in the context of quality assessment.

Secondly, the characteristics of authentic distortions are dramatically different from their synthetic counterparts, as detailed in Table \ref{tab:database_summary}. Authentic image distortions often exhibit distinct features such as uneven brightness, variations in visibility, color impairment, and diverse types of noise, \textit{etc}. Consequently, existing BIQA methods struggle to achieve satisfactory performance when applied to authentic distortions.

In light of these reasons, developing distortion-specific BIQAs for real-world distorted images are  imperative to address the challenges posed by complex and authentic artifacts. Such advancements will pave the way for more accurate and comprehensive visual quality assessment in practical applications.

\section{Deep-learned BIQA Methods}
Deep-learned BIQA methods \cite{wang2022super} have emerged as a powerful approach that directly learns quality features from distorted images in an end-to-end manner. In contrast to hand-crafted BIQAs, these methods leverage deep neural networks to automatically optimize quality forecasting models, resulting in promising performance  \cite{athar2019comprehensive}. 
Deep-learned BIQAs can be generally divided into 1) supervised learning-based and 2) unsupervised learning-based methods. Supervised methods require annotated training data, whereas unsupervised learning-based methods do not rely on quality labels during training. It is worth noting that there are other learning types, such as reinforcement learning \cite{gu2019no, saeed2022image}, that can be employed in the deep-learned BIQA task. However, these learning types are relatively less commonly used compared to supervised and unsupervised BIQA methods.

\subsection{Supervised Learning-based BIQAs}
In supervised learning-based BIQA \cite{liu2023multiscale}, the primary learning objective is to minimize the discrepancy  between the predicted score and the subjective MOS value provided by human observers. The development of supervised learning has significantly advanced the field of BIQAs.  Existing supervised learning-based BIQAs tackle the challenge of limited training samples by leveraging specific strategies, and they can be broadly categorized into two main types: 1) sample-based BIQAs and 2) constraint-based BIQAs.

\subsubsection{\textbf{Sampled-based}}
Sample-based BIQAs are mainly based on expanding the capacity of training samples, which usually utilize patch-level quality features to predict an image-level score. Existing sample-based BIQAs mainly consists of 1) annotation allocation-based and 2) annotation generation-based methods.

\noindent\textbf{Allocation-based}.
Allocation-based BIQAs directly share the same image-level annotations to all patches within a given image \cite{kang2014convolutional}. These methods have initially established a correlation between patch-level features and the overall image-level quality scores. Subsequent improvements have focused on refining this correlation by incorporating weighted features \cite{xu2016blind}, weighted decisions \cite{bosse2017deep}, and voting decisions \cite{ma2018end}. More recent attempts have extended the input image into multi-scale patches and learned a general feature representation \cite{ke2021musiq}. However, the inherent uncertainty in image content poses challenges for these allocation-based methods  to capture highly non-stationary feature representations \cite{ma2018end}. By developing more robust feature representations that can adapt to the local variations and complicated correlations between the content and distortion, allocation-based BIQAs can enhance their performance and provide more accurate quality predictions.

\noindent\textbf{Generation-based}.
Generation-based BIQAs usually learn a supervised model via patch-level scores. In the early stages, these methods have faced the challenge of lacking a region-level MOS database.  To overcome this limitation,  early methods have employed full-reference models to generate quality scores for image patches  \cite{kim2017fully}.  However, the forecasting performance is highly dependent on the adaptability and correlation between the full-reference model and the target BIQA task  \cite{zhang2020blind}. 
Recent progress has been greatly facilitated by substantial human involvement in building new databases with image and patch-level scoring labels \cite{ying2020patches}. This has allowed for better training of BIQA models by providing more accurate and reliable annotations. 
However, it is worth noting that establishing a large enough IQA database for practical quality inspection applications poses a significant cost challenge. The manual process of collecting and annotating large-scale datasets with quality labels can be time-consuming, labor-intensive, and expensive.

To sum up, existing sample-based BIQAs mainly expand training samples based on a single image modality. While this strategy aims to address the problem of limited sample annotations,  a potential drawback is the introduction of `sick' label augmentation, which can introduce noises and ultimately reduce the prediction accuracy and generalization ability of a learned model \cite{ding2020image}. In the future,  the training data volume problem can be explored by the association between different modalities.  The expansion of data modalities helps a deep-learned model to enrich low-level embedding features from different perspectives, thereby improving the robustness of the forecasting performance \cite{baltruvsaitis2018multimodal}.

\subsubsection{\textbf{Constraint-based}} 
Constraint-based BIQA methods optimize multiple loss functions simultaneously within a supervised learning framework. These methods can be roughly divided into 1) multi-task and 2) multi-objective BIQAs.

\noindent\textbf{Multi-task-based}.
Multi-task based BIQAs aim to train additional goals that are highly associated with image quality in addition to the primary quality prediction task. These additional goals can include the identification or estimation of specific distortion types within the image, the generation of error maps that highlight regions of image quality \cite{kim2019deep}, the classification of natural scene categories \cite{yan2019naturalness}, and the prediction of content attributes that contribute to image quality \cite{zhang2020blind}. By incorporating these additional tasks into the training process, multi-task based BIQAs can leverage the inherent correlations between these auxiliary tasks and the image quality to enhance the overall prediction performance.

\noindent\textbf{Multi-objective-based}.
Multi-objective based BIQAs aim to simultaneously optimize multiple constraints or regularization term to improve the model training \cite{su2020blindly}. These methods employ various techniques, such as  employing multiple loss functions for multi-scale supervision \cite{wu2020end}, introducing new normalization embeddings into the objective function \cite{li2020norm}, and using additional constraints to adjust initialization parameters \cite{zhu2020metaiqa}. Recent developments have explored the integration of constraints learned from other databases. For example, incremental learning has been used to measure correlation across databases \cite{ma2021remember}. Uncertainty prediction has been used to rank the fidelity across databases \cite{zhang2021uncertainty}. Continuous learning has been utilized to measure similarity across databases \cite{zhang2022continual}. By optimizing multiple constraints at the same time, multi-objective based BIQAs offer a more comprehensive and robust approach to visual quality assessment.

In summary, while pre-training can be effective in learning latent quality feature descriptions, it may not capture specific distortions that are not well represented in existing IQA databases. For instance, it is difficult to learn effective quality descriptions for low-light distortions because pre-training samples are mostly captured under normal lighting conditions. 
To address this difficulty, the future exploration can be conducted by leveraging from  cross-modalities. By utilizing different modalities, we can benefit from the complementary information they provide, enabling a more comprehensive assessment of image quality. Furthermore, the shared cross-modal knowledge \cite{radford2021learning} may help to extract more expressive quality representations.

\subsection{Unsupervised Learning-based BIQAs}
Unsupervised learning-BIQAs \cite{saha2023re} aim to extract latent embedding features without relying on ground-truth MOS labels. These unsupervised approaches measure the quality difference between training samples based on their latent features. Unlike supervised methods, unsupervised BIQAs do not require a large number of hand-crafted labels, making it more feasible in scenarios where explicit quality annotations are scarce. However, the absence of explicit objective functions in unsupervised learning poses challenges in designing effective training models. Existing unsupervised BIQAs can be broadly categorized into two main types: 1) metric-based and 2) domain-based methods.

\subsubsection{\textbf{Metric-based}} 
Metric-based BIQAs employ some widely used distance measurements, such as cosine similarity and  Wasserstein distance, to extract latent embedding features. These features are then employed to measure the difference between the current image sample and the other samples in the training database. The difference can be related to various factors, including distortion type, distortion level, and content category. Recently, various distortion descriptors have been developed to extract global embedding features, such as distortion type-based  discriminative learning \cite{madhusudana2022image}, distortion level-based contrastive learning \cite{wei2022contrastive}, and content category-based similarity learning  \cite{ou2021sdd}. In general, metric-based methods treat the training samples as mutual quality references and aim to maximize the differences in quality features between samples. By leveraging distance measurements and distortion descriptors, these methods enhance the discriminative ability of the learned embeddings and improve the accuracy of quality assessment.

\subsubsection{\textbf{Domain-based}} 
Domain-based BIQAs commonly design domain alignment constraints, and measure the quality difference between samples from different domains. These methods leverage error metrics defined in a target domain to assess the quality difference of each sample in a source domain. Early domain-based methods directly added different levels of synthetic distortion and learned to rank them \cite{liu2017rankiqa,ma2017dipiq,liu2019exploiting}. Subsequent domain adversarial methods employed a generator to construct a target domain and then guided  the source domain for adversarial learning  \cite{ren2018ran4iqa,lin2018hallucinated}.  More recently, domain adaptation methods have utilized additional large databases to construct the target domain and guide the source domain to learn the rules of quality description in the target domain \cite{chen2021unsupervised,chen2021no}. It is important to note that domain-based methods usually require strict assumptions, which make them difficult to meet the model requirements when the distortion type of a testing image is unknown.
\begin{figure*}[!t]
\centering
\includegraphics[width=0.95\textwidth]{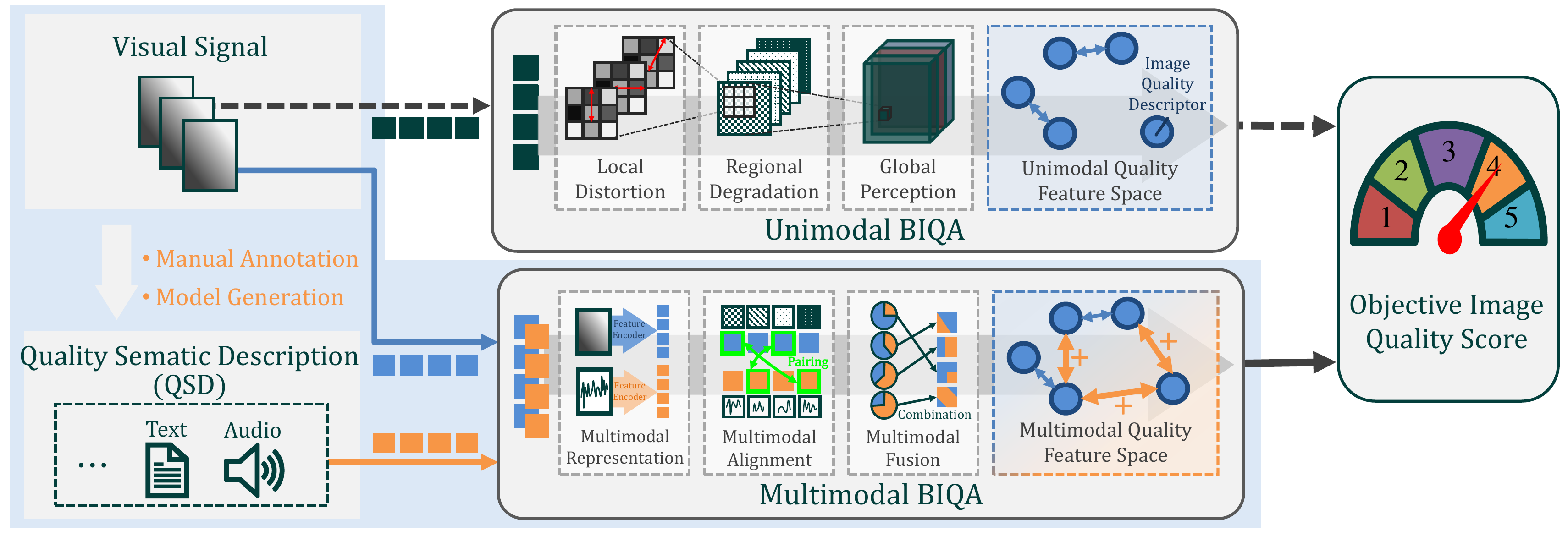}\\
\caption{\textit{Relationship between unimodal and multimodal BIQAs} \cite{wang2023blind}. Humans are better at perceiving image quality through semantic descriptions rather than quantitative values, which reveals that quality semantic description (QSD) is a very useful modality for BIQA  modeling.}
\label{fig:unimodal_multimodal}
\end{figure*}

\subsection{Summary}
Existing deep-learned BIQAs have primarily focused on a single image modality, while multimodal-driven  BIQAs are still in their early stage. However, the incorporation of multimodal information provides a new and feasible solution to the problem of insufficient training samples. One advantage of multimodal BIQA is the homology of multimodal data, which suggests that training information between different modalities can be complementary and shared, which facilitates the learning of highly descriptive quality features. This enables the learning of highly descriptive quality features that may not be fully captured by a single modality alone. Furthermore, the heterogeneity of multimodal data expands the width and depth of training information. Each modality provides unique perspectives and cues related to image quality, allowing for a more comprehensive understanding of quality assessment. Given these advantages, it is both necessary and meaningful to embark on a new exploration of multimodal approaches in the BIQA task.

\section{Multimodal Quality Assessment}
Multimodal quality assessment methods \cite{xie2023pmbqa} are currently in their early stages of development. To our knowledge, there are two primary modality combinations that have been explored: visual-audio and visual-text methods. These modalities aim to leverage the complementary information from cross sources, such as visual and auditory cues or visual and textual cues, to quantify the overall quality of multimedia content.

\subsection{Visual-audio Quality Assessment}
Visual-audio quality assessment \cite{min2020study} refers to the quantitative evaluation of user experience,  involving the quality assessment across multiple media modalities. It can be further divided into two main types: degradation-based methods and perception-based methods. By jointly considering the quality of both visual and audio modalities, visual-audio quality assessment provides a comprehensive evaluation of the overall user experience in multimedia applications.

\subsubsection{\textbf{Degradation-based}}
Existing degradation-based methods commonly measure the user-perceived quality score of visual and audio independently, and then employ a combination rule to predict the overall score.
This paradigm assumes that the degradation of video and audio is solely influenced by factors such as signal capture equipment \cite{porikli2011multimedia}, compression, transmission, and terminals.  It also assumes that the degradation of one modality does not directly impact on that of the other (\textit{i.e.}, video degradation does not cause audio degradation). The combination rule used in these methods mainly involve addition \cite{hands2004basic}, multiplication \cite{winkler2006perceived}, vote \cite{waltl2010improving}, weighted Minkowski \cite{martinez2018combining}, \textit{etc}. However, these degradation-based approaches overlook the potential interaction and dependencies between cross-media information. As a result, simply combining individual quality scores may not accurately measure the user experience score.

\subsubsection{\textbf{Perception-based}}
The perception-based methods aim to learn a joint perceptual feature space, where video and audio features can both be represented and combined to predict the overall quality score. This paradigm is  motivated by psychophysical experiments, such as the McGurk effect \cite{mcgurk1976hearing}, where cross-media perceptual interaction has a greater impact on the user-perceived quality. These experiments reveal that the user-perceived video quality is more favorable when paired with high-fidelity sound than when used alone. Conversely, the perception of audio quality is more favorable when assessed alone, rather than when paired with high-quality videos \cite{storms2000interactions}. 
Therefore, the user-perceived quality should be considered based on the combination of perceptual features to capture the above cross-media biases. Existing explorations mainly focus on the  combination of spatial perception \cite{cao2021deep}, temporal perception \cite{ying2022telepresence}, and spatio-temporal perception \cite{min2020study}. By incorporating these perceptual features and capturing their interactions, these methods enable more reliable assessments of the user experience in multimedia applications.

\subsection{Visual-text Quality Assessment}
Recently, a multimodal low-light image quality (MLIQ) \cite{wang2023blind} database has been constructed, including the image and text modalities. The text modality mainly consists of semantic descriptions  that capture the visual quality of each image sample. As discussed in \cite{min2020study,pinson2012influence}, there are several multimodal quality assessment databases separately established for video and audio, respectively. Among these visual-audio databases, the closest to the MLIQ is the LIVE-SJTU audio and video quality assessment database (A/V-QA) \cite{min2020study}.

Unlike the A/V-QA, MLIQ consists of image samples and their text modalities: 1) In addition to visual modality, the text modality in the MLIQ reflects the image quality in terms of semantic visual descriptions, while that the audio modality in A/V-QA can be silence, noise, pure music, pure speech, or speech over background sound. 2) MLIQ contains various types of authentic distortions, while A/V-QA only contains synthetic compression and sampling distortions. 3) MLIQ is specifically established to study the impact of auxiliary modalities on the multimodal-based  IQA task. In contrast,  A/V-QA is constructed to study the impact of both visual and audio distortion on the overall user experience. This distinction implies that the MOS labels of MLIQ solely depend on the image modality, while these of A/V-QA depend on both the visual and audio modalities.

Based on the MLIQ database, they have developed a blind multimodal quality assessment (BMQA) method to integrate cross-modal features. This method consists of several learning modules, including multimodal quality representation, latent feature alignment, and fusion prediction. To improve the efficiency of deep model training, they have also employed an effective BMQA method by incorporating both multimodal self-supervision and supervision.  The information integration between visual and text modalities is beneficial to expanding the breadth of quality information available, allowing for a more robust and comprehensive evaluation of image quality. By incorporating cross-modal features, the BMQA method demonstrates superior generalization ability, enabling accurate prediction performance even on unseen or unfamiliar data.

\subsection{Summary}
It is worth noting that humans are better at measuring image quality by semantic description rather than quantitative score \cite{yang2022fine}, and hence the text-based quality descriptions can be a very useful modality in the modeling of BMQA. By incorporating text-based information, BMQAs can simulate the inherent ability of humans to capture and represent visual quality in a more comprehensive manner. The relationship between unimodal and multimodal BIQA is illustrated in Fig. \ref{fig:unimodal_multimodal}, which provides a visual representation of how the different modalities, such as image, audio, and text, interact and contribute to the overall quality assessment process.

In addition, the differences between existing visual-audio and visual-text quality indicators are obvious. Firstly, the former is designed to forecast the joint quality score of the user experience across different media types, while the latter is focused on predicting the visual quality of a given image.  Secondly, in video-audio methods, the audio modality is usually unavailable in the BIQA task, and they become inapplicable. In the visual-text methods, the text modality can be easily generated by image captioning \cite{vinyals2016show} or large language models (LLMs) \cite{liu2023pre}, and the lack of the text modality is no longer a problem.

Finally,  multimodality-driven BIQAs are currently in their nascent phase, but BMQA provides a new and promising perspective. On one hand, the homogeneity of multimodal data shows that training information can be supplementary or shared, which facilitates the learning of highly descriptive quality features. On the other hand, the heterogeneity of multimodal data can expand the width and depth of training information, which is expected to improve the forecasting performance. To this end, it is highly desired to develop a multimodal quality indicator.

\section{Conclusion and Future Work}
This survey has provided a detailed analysis and discussion of recent progress in BIQAs. We have examined various aspects of BIQAs, including hand-crafted methods, deep-learned approaches, multimodal quality assessment, and representative databases. Through this comprehensive overview, we have outlined the developments and highlighted the challenges that still remain in the field of BIQA. Specifically, hand-crafted BIQA methods have evolved to address distortion-specific and general-purpose images. Deep-learned BIQA methods, which utilize supervised and unsupervised learning techniques, have shown promising results in capturing complex image quality attributes. The integration of multimodal information, such as visual-audio and visual-text interactions, has emerged as a valuable direction to improving the accuracy and robustness of BIQA models. Additionally, the use of databases containing synthetic and authentic distortions has played a crucial role in training and evaluating these models.

Despite the promising progress made, there are several challenges that still need to be addressed in the region of BIQA. These challenges include handling diverse real-world conditions, such as low-light environments and immersive displays, as well as incorporating user preferences and subjective assessments. Future research should focus on developing more efficient and accurate BIQA models capable of addressing these challenges and providing reliable application-specific quality assessments. We believe that this new survey serves as a valuable resource for academic researchers and industry professionals, offering insights into the development of more accurate and reliable BIQAs.


{
\balance
\bibliographystyle{ieee_fullname}
\bibliographystyle{IEEEtran}
\bibliography{iqa2024bib}

\begin{thebibliography}{100}\itemsep=-1pt

\bibitem{amini2023towards}
Reza Amini~Gougeh and Tiago~H Falk.
\newblock Towards instrumental quality assessment of multisensory immersive
  experiences using a biosensor-equipped head-mounted display.
\newblock {\em Springer Quality and User Experience}, 8(1):9, 2023.

\bibitem{athar2019comprehensive}
Shahrukh Athar and Zhou Wang.
\newblock A comprehensive performance evaluation of image quality assessment
  algorithms.
\newblock {\em IEEE Access}, 7:140030--140070, 2019.

\bibitem{bai2021blind}
Yongqiang Bai, Zhongjie Zhu, Gangyi Jiang, and Huifang Sun.
\newblock {Blind quality assessment of screen content images via macro-micro
  modeling of tensor domain dictionary}.
\newblock {\em IEEE Transactions on Multimedia}, 23(1):4259--4271, 2021.

\bibitem{baltruvsaitis2018multimodal}
Tadas Baltru{\v{s}}aitis, Chaitanya Ahuja, and Louis-Philippe Morency.
\newblock {Multimodal machine learning: A survey and taxonomy}.
\newblock {\em IEEE Transactions on Pattern Analysis and Machine Intelligence},
  41(2):423--443, 2018.

\bibitem{bosse2017deep}
Sebastian Bosse, Dominique Maniry, Klaus-Robert M{\"u}ller, Thomas Wiegand, and
  Wojciech Samek.
\newblock {Deep neural networks for no-reference and full-reference image
  quality assessment}.
\newblock {\em IEEE Transactions on Image Processing}, 27(1):206--219, 2018.

\bibitem{cai2023blind}
Rongtai Cai and Ming Fang.
\newblock Blind image quality assessment by simulating the visual cortex.
\newblock {\em Springer The Visual Computer}, 39(10):4639--4656, 2023.

\bibitem{cao2021deep}
Yuqin Cao, Xiongkuo Min, Wei Sun, and Guangtao Zhai.
\newblock {Deep Neural Networks For Full-Reference And No-Reference
  Audio-Visual Quality Assessment}.
\newblock In {\em IEEE International Conference on Image Processing (ICIP)},
  pages 1429--1433, 2021.

\bibitem{cao2023subjective}
Yuqin Cao, Xiongkuo Min, Wei Sun, and Guangtao Zhai.
\newblock {Subjective and objective audio-visual quality assessment for user
  generated content}.
\newblock {\em IEEE Transactions on Image Processing}, 32:3847--3861, 2023.

\bibitem{chen2021no}
Baoliang Chen, Haoliang Li, Hongfei Fan, and Shiqi Wang.
\newblock {No-reference screen content image quality assessment with
  unsupervised domain adaptation}.
\newblock {\em IEEE Transactions on Image Processing}, 30(1):5463--5476, 2021.

\bibitem{chen2020study}
Meixu Chen, Yize Jin, Todd Goodall, Xiangxu Yu, and Alan~Conrad Bovik.
\newblock {Study of 3D virtual reality picture quality}.
\newblock {\em IEEE Journal of Selected Topics in Signal Processing},
  14(1):89--102, 2020.

\bibitem{chen2021unsupervised}
Pengfei Chen, Leida Li, Jinjian Wu, Weisheng Dong, and Guangming Shi.
\newblock {Unsupervised curriculum domain adaptation for no-reference video
  quality assessment}.
\newblock In {\em IEEE International Conference on Computer Vision (ICCV)},
  pages 5178--5187, 2021.

\bibitem{deng2017image}
Yubin Deng, Chen~Change Loy, and Xiaoou Tang.
\newblock {Image aesthetic assessment: An experimental survey}.
\newblock {\em IEEE Signal Processing Magazine}, 34(4):80--106, 2017.

\bibitem{ding2020image}
Keyan Ding, Kede Ma, Shiqi Wang, and Eero~P. Simoncelli.
\newblock {Image quality assessment: Unifying structure and texture
  similarity}.
\newblock {\em IEEE Transactions on Pattern Analysis and Machine Intelligence},
  44(5):2567--2581, 2022.

\bibitem{dost2022reduced}
Shahi Dost, Faryal Saud, Maham Shabbir, Muhammad~Gufran Khan, Muhammad Shahid,
  and Benny Lovstrom.
\newblock {Reduced reference image and video quality assessments: review of
  methods}.
\newblock {\em EURASIP Journal on Image and Video Processing}, 2022(1):1--31,
  2022.

\bibitem{duan2018perceptual}
Huiyu Duan, Guangtao Zhai, Xiongkuo Min, Yucheng Zhu, Yi Fang, and Xiaokang
  Yang.
\newblock {Perceptual Quality Assessment of Omnidirectional Images}.
\newblock In {\em IEEE International Symposium on Circuits and Systems
  (ISCAS)}, pages 1--5, 2018.

\bibitem{erlenbusch2023thermal}
Fabian Erlenbusch, Constanze Merkt, Bernardo de Oliveira, Alexander Gatter,
  Friedhelm Schwenker, Ulrich Klauck, and Michael Teutsch.
\newblock {Thermal Infrared Single Image Dehazing and Blind Image Quality
  Assessment}.
\newblock In {\em IEEE Conference on Computer Vision and Pattern Recognition
  (CVPR)}, pages 459--469, 2023.

\bibitem{fang2020perceptual}
Yuming Fang, Rengang Du, Yifan Zuo, Wenying Wen, and Leida Li.
\newblock {Perceptual quality assessment for screen content images by spatial
  continuity}.
\newblock {\em IEEE Transactions on Circuits and Systems for Video Technology},
  30(11):4050--4063, 2020.

\bibitem{fang2022perceptual}
Yuming Fang, Liping Huang, Jiebin Yan, Xuelin Liu, and Yang Liu.
\newblock {Perceptual quality assessment of omnidirectional images}.
\newblock In {\em AAAI Conference on Artificial Intelligence (AAAI)}, pages
  580--588, 2022.

\bibitem{fang2017no}
Yuming Fang, Jiebin Yan, Leida Li, Jinjian Wu, and Weisi Lin.
\newblock {No reference quality assessment for screen content images with both
  local and global feature representation}.
\newblock {\em IEEE Transactions on Image Processing}, 27(4):1600--1610, 2017.

\bibitem{freitas2018no}
Pedro~Garcia Freitas, Welington~YL Akamine, and Mylene~CQ Farias.
\newblock {No-reference image quality assessment using orthogonal color planes
  patterns}.
\newblock {\em IEEE Transactions on Multimedia}, 20(12):3353--3360, 2018.

\bibitem{gu2019no}
Jie Gu, Gaofeng Meng, Cheng Da, Shiming Xiang, and Chunhong Pan.
\newblock {No-reference image quality assessment with reinforcement recursive
  list-wise ranking}.
\newblock In {\em AAAI Conference on Artificial Intelligence (AAAI)}, pages
  8336--8343, 2019.

\bibitem{gu2017no}
Ke Gu, Jun Zhou, Jun-Fei Qiao, Guangtao Zhai, Weisi Lin, and Alan~Conrad Bovik.
\newblock {No-reference quality assessment of screen content pictures}.
\newblock {\em IEEE Transactions on Image Processing}, 26(8):4005--4018, 2017.

\bibitem{guo2019peid}
Shangwei Guo, Tao Xiang, Xiaoguo Li, and Ying Yang.
\newblock {PEID: A perceptually encrypted image database for visual security
  evaluation}.
\newblock {\em IEEE Transactions on Information Forensics and Security},
  15(1):1151--1163, 2019.

\bibitem{hands2004basic}
David~S Hands.
\newblock {A basic multimedia quality model}.
\newblock {\em IEEE Transactions on Multimedia}, 6(6):806--816, 2004.

\bibitem{hosu2020koniq}
Vlad Hosu, Hanhe Lin, Tamas Sziranyi, and Dietmar Saupe.
\newblock {KonIQ-10k: An ecologically valid database for deep learning of blind
  image quality assessment}.
\newblock {\em IEEE Transactions on Image Processing}, 29(1):4041--4056, 2020.

\bibitem{hou2023uid2021}
Guojia Hou, Yuxuan Li, Huan Yang, Kunqian Li, and Zhenkuan Pan.
\newblock {UID2021: an underwater image dataset for evaluation of no-reference
  quality assessment metrics}.
\newblock {\em ACM Transactions on Multimedia Computing, Communications and
  Applications}, 19(4):1--24, 2023.

\bibitem{huang2018modeling}
Mingkai Huang, Qiu Shen, Zhan Ma, Alan~Conrad Bovik, Praful Gupta, Rongbing
  Zhou, and Xun Cao.
\newblock {Modeling the perceptual quality of immersive images rendered on head
  mounted displays: Resolution and compression}.
\newblock {\em IEEE Transactions on Image Processing}, 27(12):6039--6050, 2018.

\bibitem{jiang2021cubemap}
Hao Jiang, Gangyi Jiang, Mei Yu, Yun Zhang, You Yang, Zongju Peng, Fen Chen,
  and Qingbo Zhang.
\newblock {Cubemap-based perception-driven blind quality assessment for
  360-degree images}.
\newblock {\em IEEE Transactions on Image Processing}, 30(1):2364--2377, 2021.

\bibitem{jiang2021blind}
Mingxing Jiang, Liquan Shen, Min Hu, Ping An, and Fuji Ren.
\newblock {Blind Quality Evaluator of Tone-Mapped HDR and Multi-Exposure Fused
  Images for Electronic Display}.
\newblock {\em IEEE Transactions on Consumer Electronics}, 67(4):350--362,
  2021.

\bibitem{kang2014convolutional}
Le Kang, Peng Ye, Yi Li, and David Doermann.
\newblock {Convolutional neural networks for no-reference image quality
  assessment}.
\newblock In {\em IEEE Conference on Computer Vision and Pattern Recognition
  (CVPR)}, pages 1733--1740, 2014.

\bibitem{ke2021musiq}
Junjie Ke, Qifei Wang, Yilin Wang, Peyman Milanfar, and Feng Yang.
\newblock {MUSIQ: Multi-scale image quality transformer}.
\newblock In {\em IEEE International Conference on Computer Vision (ICCV)},
  pages 5148--5157, 2021.

\bibitem{kim2017fully}
Jongyoo Kim and Sanghoon Lee.
\newblock {Fully deep blind image quality predictor}.
\newblock {\em IEEE Journal of Selected Topics in Signal Processing},
  11(1):206--220, 2017.

\bibitem{kim2019deep}
Jongyoo Kim, Anh-Duc Nguyen, and Sanghoon Lee.
\newblock {Deep CNN-based blind image quality predictor}.
\newblock {\em IEEE Transactions on Neural Networks and Learning Systems},
  30(1):11--24, 2019.

\bibitem{korshunov2015subjective}
Pavel Korshunov, Philippe Hanhart, Thomas Richter, Alessandro Artusi, Rafa{\l}
  Mantiuk, and Touradj Ebrahimi.
\newblock {Subjective quality assessment database of HDR images compressed with
  JPEG XT}.
\newblock In {\em IEEE International Workshop on Quality of Multimedia
  Experience (QoMEX)}, pages 1--6, 2015.

\bibitem{kundu2017large}
Debarati Kundu, Deepti Ghadiyaram, Alan~C Bovik, and Brian~L Evans.
\newblock {Large-scale crowdsourced study for tone-mapped HDR pictures}.
\newblock {\em IEEE Transactions on Image Processing}, 26(10):4725--4740, 2017.

\bibitem{kundu2017no}
Debarati Kundu, Deepti Ghadiyaram, Alan~C Bovik, and Brian~L Evans.
\newblock {No-reference quality assessment of tone-mapped HDR pictures}.
\newblock {\em IEEE Transactions on Image Processing}, 26(6):2957--2971, 2017.

\bibitem{larson2009categorical}
EC Larson and DM Chandler.
\newblock {Categorical subjective image quality CSIQ database}, 2009.

\bibitem{li2020norm}
Dingquan Li, Tingting Jiang, and Ming Jiang.
\newblock {Norm-in-norm loss with faster convergence and better performance for
  image quality assessment}.
\newblock In {\em ACM International Conference on Multimedia (MM)}, pages
  789--797, 2020.

\bibitem{li2017bsd}
Qiaohong Li, Weisi Lin, and Yuming Fang.
\newblock {BSD: Blind image quality assessment based on structural
  degradation}.
\newblock {\em Elsevier Neurocomputing}, 236(1):93--103, 2017.

\bibitem{li2016blind}
Qiaohong Li, Weisi Lin, Jingtao Xu, and Yuming Fang.
\newblock {Blind image quality assessment using statistical structural and
  luminance features}.
\newblock {\em IEEE Transactions on Multimedia}, 18(12):2457--2469, 2016.

\bibitem{lin2019kadid}
Hanhe Lin, Vlad Hosu, and Dietmar Saupe.
\newblock {KADID-10k: A large-scale artificially distorted IQA database}.
\newblock In {\em International Conference on Quality of Multimedia Experience
  (QoMEX)}, pages 1--3, 2019.

\bibitem{lin2018hallucinated}
Kwan-Yee Lin and Guanxiang Wang.
\newblock {Hallucinated-IQA: No-reference image quality assessment via
  adversarial learning}.
\newblock In {\em IEEE Conference on Computer Vision and Pattern Recognition
  (CVPR)}, pages 732--741, 2018.

\bibitem{liu2023multiscale}
Manni Liu, Jiabin Huang, Delu Zeng, Xinghao Ding, and John Paisley.
\newblock {A Multiscale Approach to Deep Blind Image Quality Assessment}.
\newblock {\em IEEE Transactions on Image Processing}, 32:1656--1667, 2023.

\bibitem{liu2023pre}
Pengfei Liu, Weizhe Yuan, Jinlan Fu, Zhengbao Jiang, Hiroaki Hayashi, and
  Graham Neubig.
\newblock Pre-train, prompt, and predict: A systematic survey of prompting
  methods in natural language processing.
\newblock {\em ACM Computing Surveys}, 55(9):1--35, 2023.

\bibitem{liu2018no}
Tsung-Jung Liu and Kuan-Hsien Liu.
\newblock {No-reference image quality assessment by wide-perceptual-domain
  scorer ensemble method}.
\newblock {\em IEEE Transactions on Image Processing}, 27(3):1138--1151, 2018.

\bibitem{liu2017rankiqa}
Xialei Liu, Joost Van De~Weijer, and Andrew~D. Bagdanov.
\newblock {RankIQA: Learning from rankings for no-reference image quality
  assessment}.
\newblock In {\em IEEE International Conference on Computer Vision (ICCV)},
  pages 1040--1049, 2017.

\bibitem{liu2019exploiting}
Xialei Liu, Joost Van De~Weijer, and Andrew~D Bagdanov.
\newblock {Exploiting unlabeled data in CNNs by self-supervised learning to
  rank}.
\newblock {\em IEEE Transactions on Pattern Analysis and Machine Intelligence},
  41(8):1862--1878, 2019.

\bibitem{liu2020blind}
Yutao Liu, Ke Gu, Xiu Li, and Yongbing Zhang.
\newblock {Blind image quality assessment by natural scene statistics and
  perceptual characteristics}.
\newblock {\em ACM Transactions on Multimedia Computing, Communications, and
  Applications}, 16(3):1--91, 2020.

\bibitem{ma2016waterloo}
Kede Ma, Zhengfang Duanmu, Qingbo Wu, Zhou Wang, Hongwei Yong, Hongliang Li,
  and Lei Zhang.
\newblock {Waterloo exploration database: New challenges for image quality
  assessment models}.
\newblock {\em IEEE Transactions on Image Processing}, 26(2):1004--1016, 2016.

\bibitem{ma2017dipiq}
Kede Ma, Wentao Liu, Tongliang Liu, Zhou Wang, and Dacheng Tao.
\newblock {dipIQ: Blind image quality assessment by learning-to-rank
  discriminable image pairs}.
\newblock {\em IEEE Transactions on Image Processing}, 26(8):3951--3964, 2017.

\bibitem{ma2018end}
Kede Ma, Wentao Liu, Kai Zhang, Zhengfang Duanmu, Zhou Wang, and Wangmeng Zuo.
\newblock {End-to-end blind image quality assessment using deep neural
  networks}.
\newblock {\em IEEE Transactions on Image Processing}, 27(3):1202--1213, 2018.

\bibitem{ma2021remember}
Rui Ma, Hanxiao Luo, Qingbo Wu, King~Ngi Ngan, Hongliang Li, Fanman Meng, and
  Linfeng Xu.
\newblock {Remember and Reuse: Cross-Task Blind Image Quality Assessment via
  Relevance-aware Incremental Learning}.
\newblock In {\em ACM International Conference on Multimedia (MM)}, pages
  5248--5256, 2021.

\bibitem{madhusudana2022image}
Pavan~C Madhusudana, Neil Birkbeck, Yilin Wang, Balu Adsumilli, and Alan~C
  Bovik.
\newblock {Image quality assessment using contrastive learning}.
\newblock {\em IEEE Transactions on Image Processing}, 31(1):4149--4161, 2022.

\bibitem{madhusudana2019subjective}
Pavan~Chennagiri Madhusudana and Rajiv Soundararajan.
\newblock {Subjective and objective quality assessment of stitched images for
  virtual reality}.
\newblock {\em IEEE Transactions on Image Processing}, 28(11):5620--5635, 2019.

\bibitem{saeed2020multi}
Saeed Mahmoudpour and Peter Schelkens.
\newblock {A Multi-Attribute Blind Quality Evaluator for Tone-Mapped Images}.
\newblock {\em IEEE Transactions on Multimedia}, 22(8):1939--1954, 2020.

\bibitem{manap2015non}
Redzuan~Abdul Manap and Ling Shao.
\newblock {Non-distortion-specific no-reference image quality assessment: A
  survey}.
\newblock {\em Elsevier Information Sciences}, 301:141--160, 2015.

\bibitem{martinez2018combining}
Helard A~Becerra Martinez and Mylene~CQ Farias.
\newblock {Combining audio and video metrics to assess audio-visual quality}.
\newblock {\em Springer Multimedia Tools and Applications},
  77(18):23993--24012, 2018.

\bibitem{mcgurk1976hearing}
Harry McGurk and John MacDonald.
\newblock {Hearing lips and seeing voices}.
\newblock {\em Nature}, 264(5588):746--748, 1976.

\bibitem{min2021screen}
Xiongkuo Min, Ke Gu, Guangtao Zhai, Xiaokang Yang, Wenjun Zhang, Patrick
  Le~Callet, and Chang~Wen Chen.
\newblock {Screen content quality assessment: Overview, benchmark, and beyond}.
\newblock {\em ACM Computing Surveys}, 54(9):1--36, 2021.

\bibitem{min2017unified}
Xiongkuo Min, Kede Ma, Ke Gu, Guangtao Zhai, Zhou Wang, and Weisi Lin.
\newblock {Unified blind quality assessment of compressed natural, graphic, and
  screen content images}.
\newblock {\em IEEE Transactions on Image Processing}, 26(11):5462--5474, 2017.

\bibitem{min2020study}
Xiongkuo Min, Guangtao Zhai, Jiantao Zhou, Mylene~CQ Farias, and Alan~Conrad
  Bovik.
\newblock {Study of subjective and objective quality assessment of audio-visual
  signals}.
\newblock {\em IEEE Transactions on Image Processing}, 29(5588):6054--6068,
  2020.

\bibitem{min2020multimodal}
Xiongkuo Min, Guangtao Zhai, Jiantao Zhou, Xiao-Ping Zhang, Xiaokang Yang, and
  Xinping Guan.
\newblock {A multimodal saliency model for videos with high audio-visual
  correspondence}.
\newblock {\em IEEE Transactions on Image Processing}, 29:3805--3819, 2020.

\bibitem{mittal2012no}
Anish Mittal, Anush~Krishna Moorthy, and Alan~Conrad Bovik.
\newblock {No-reference image quality assessment in the spatial domain}.
\newblock {\em IEEE Transactions on Image Processing}, 21(12):4695--4708, 2012.

\bibitem{mittal2013making}
Anish Mittal, Rajiv Soundararajan, and Alan~C. Bovik.
\newblock {Making a `Completely Blind' Image Quality Analyzer}.
\newblock {\em IEEE Signal Processing Letters}, 20(3):209--212, 2013.

\bibitem{moorthy2010two}
Anush~Krishna Moorthy and Alan~Conrad Bovik.
\newblock {A two-step framework for constructing blind image quality indices}.
\newblock {\em IEEE Signal Processing Letters}, 17(5):513--516, 2010.

\bibitem{moorthy2011blind}
Anush~Krishna Moorthy and Alan~Conrad Bovik.
\newblock {Blind image quality assessment: From natural scene statistics to
  perceptual quality}.
\newblock {\em IEEE Transactions on Image Processing}, 20(12):3350--3364, 2011.

\bibitem{ni2017esim}
Zhangkai Ni, Lin Ma, Huanqiang Zeng, Jing Chen, Canhui Cai, and Kai-Kuang Ma.
\newblock {ESIM: Edge similarity for screen content image quality assessment}.
\newblock {\em IEEE Transactions on Image Processing}, 26(10):4818--4831, 2017.

\bibitem{ou2021sdd}
Fu-Zhao Ou, Xingyu Chen, Ruixin Zhang, Yuge Huang, Shaoxin Li, Jilin Li, Yong
  Li, Liujuan Cao, and Yuan-Gen Wang.
\newblock {Sdd-fiqa: Unsupervised face image quality assessment with similarity
  distribution distance}.
\newblock In {\em IEEE Conference on Computer Vision and Pattern Recognition
  (CVPR)}, pages 7670--7679, 2021.

\bibitem{pinson2012influence}
Margaret~H Pinson, Lucjan Janowski, Romuald P{\'e}pion, Quan Huynh-Thu,
  Christian Schmidmer, Phillip Corriveau, Audrey Younkin, Patrick Le~Callet,
  Marcus Barkowsky, and William Ingram.
\newblock {The influence of subjects and environment on audiovisual subjective
  tests: An international study}.
\newblock {\em IEEE Journal of Selected Topics in Signal Processing},
  6(6):640--651, 2012.

\bibitem{ponomarenko2015image}
Nikolay Ponomarenko, Lina Jin, Oleg Ieremeiev, Vladimir Lukin, Karen
  Egiazarian, Jaakko Astola, Benoit Vozel, Kacem Chehdi, Marco Carli, Federica
  Battisti, et~al.
\newblock {Image database TID2013: Peculiarities, results and perspectives}.
\newblock {\em Signal processing: Image communication}, 30:57--77, 2015.

\bibitem{ponomarenko2009tid2008}
Nikolay Ponomarenko, Vladimir Lukin, Alexander Zelensky, Karen Egiazarian,
  Marco Carli, and Federica Battisti.
\newblock {TID2008 - {A} database for evaluation of full-reference visual
  quality assessment metrics}.
\newblock {\em Advances of Modern Radioelectronics}, 10(4):30--45, 2009.

\bibitem{porikli2011multimedia}
Fatih Porikli, Al Bovik, Chris Plack, Ghassan AlRegib, Joyce Farrell, Patrick
  Le~Callet, Quan Huynh-Thu, Sebastian M{\"o}ller, and Stefan Winkler.
\newblock {Multimedia quality assessment [DSP Forum]}.
\newblock {\em IEEE Signal Processing Magazine}, 28(6):164--177, 2011.

\bibitem{radford2021learning}
Alec Radford, Jong~Wook Kim, Chris Hallacy, Aditya Ramesh, Gabriel Goh,
  Sandhini Agarwal, Girish Sastry, Amanda Askell, Pamela Mishkin, Jack Clark,
  et~al.
\newblock {Learning transferable visual models from natural language
  supervision}.
\newblock In {\em PMLR International Conference on Machine Learning (ICML)},
  pages 8748--8763, 2021.

\bibitem{ren2018ran4iqa}
Hongyu Ren, Diqi Chen, and Yizhou Wang.
\newblock {RAN4IQA: Restorative adversarial nets for no-reference image quality
  assessment}.
\newblock In {\em AAAI Conference on Artificial Intelligence (AAAI)}, pages
  7308--7314, 2018.

\bibitem{rousselot2018impacts}
Maxime Rousselot, {\'E}ric Auffret, Xavier Ducloux, Olivier Le~Meur, and
  R{\'e}mi Cozot.
\newblock {Impacts of viewing conditions on hdr-vdp2}.
\newblock In {\em IEEE European Signal Processing Conference (EUSIPCO)}, pages
  1442--1446, 2018.

\bibitem{saad2012blind}
Michele~A Saad, Alan~C Bovik, and Christophe Charrier.
\newblock {Blind image quality assessment: A natural scene statistics approach
  in the DCT domain}.
\newblock {\em IEEE Transactions on Image Processing}, 21(8):3339--3352, 2012.

\bibitem{saeed2022image}
Shaheer~U. Saeed, Yunguan Fu, Vasilis Stavrinides, Zachary~M.C. Baum, Qianye
  Yang, Mirabela Rusu, Richard~E. Fan, Geoffrey~A. Sonn, J.~Alison Noble,
  Dean~C. Barratt, and Yipeng Hu.
\newblock {Image quality assessment for machine learning tasks using
  meta-reinforcement learning}.
\newblock {\em Elsevier Medical Image Analysis}, 78(1):102427, 2022.

\bibitem{saha2023re}
Avinab Saha, Sandeep Mishra, and Alan~C Bovik.
\newblock {Re-IQA: Unsupervised Learning for Image Quality Assessment in the
  Wild}.
\newblock In {\em IEEE Conference on Computer Vision and Pattern Recognition
  (CVPR)}, pages 5846--5855, 2023.

\bibitem{sheikh2006image}
Hamid~R Sheikh and Alan~C Bovik.
\newblock {Image information and visual quality}.
\newblock {\em IEEE Transactions on image processing}, 15(2):430--444, 2006.

\bibitem{shi2022objective}
Ran Shi, Jing Ma, King~Ngi Ngan, Jian Xiong, and Tong Qiao.
\newblock {Objective Object Segmentation Visual Quality Evaluation: Quality
  Measure and Pooling Method}.
\newblock {\em ACM Transactions on Multimedia Computing, Communications, and
  Applications}, 18(3):1--19, 2022.

\bibitem{shi2015study}
Sheng Shi, Xiang Zhang, Shiqi Wang, Ruiqin Xiong, and Siwei Ma.
\newblock {Study on subjective quality assessment of screen content images}.
\newblock In {\em IEEE Picture Coding Symposium (PCS)}, pages 75--79, 2015.

\bibitem{song2019harmonized}
Guoli Song, Shuhui Wang, Qingming Huang, and Qi Tian.
\newblock {Harmonized multimodal learning with Gaussian process latent variable
  models}.
\newblock {\em IEEE Transactions on Pattern Analysis and Machine Intelligence},
  43(3):858--872, 2019.

\bibitem{stkepien2022brief}
Igor Stepien and Mariusz Oszust.
\newblock {A brief survey on no-reference image quality assessment methods for
  magnetic resonance images}.
\newblock {\em MDPI Journal of Imaging}, 8(6):160, 2022.

\bibitem{storms2000interactions}
Russell~L Storms and Michael~J Zyda.
\newblock {Interactions in perceived quality of auditory-visual displays}.
\newblock {\em MIT Press Presence: Teleoperators \& Virtual Environments},
  9(6):557--580, 2000.

\bibitem{stutz2010subjective}
Thomas St{\"u}tz, Vinod Pankajakshan, Florent Autrusseau, Andreas Uhl, and
  Heinz Hofbauer.
\newblock {Subjective and objective quality assessment of transparently
  encrypted JPEG2000 images}.
\newblock In {\em ACM workshop on Multimedia and Security}, pages 247--252,
  2010.

\bibitem{su2020blindly}
Shaolin Su, Qingsen Yan, Yu Zhu, Cheng Zhang, Xin Ge, Jinqiu Sun, and Yanning
  Zhang.
\newblock {Blindly assess image quality in the wild guided by a self-adaptive
  hyper network}.
\newblock In {\em IEEE Conference on Computer Vision and Pattern Recognition
  (CVPR)}, pages 3667--3676, 2020.

\bibitem{sui2022perceptual}
Xiangjie Sui, Kede Ma, Yiru Yao, and Yuming Fang.
\newblock {Perceptual quality assessment of omnidirectional images as moving
  camera videos}.
\newblock {\em IEEE Transactions on Visualization and Computer Graphics},
  28(8):3022--3034, 2022.

\bibitem{sun2018large}
Wei Sun, Ke Gu, Siwei Ma, Wenhan Zhu, Ning Liu, and Guangtao Zhai.
\newblock {A large-scale compressed 360-degree spherical image database: From
  subjective quality evaluation to objective model comparison}.
\newblock In {\em IEEE International Workshop on Multimedia Signal Processing
  (MMSP)}, pages 1--6, 2018.

\bibitem{sun2023blind}
Wei Sun, Xiongkuo Min, Danyang Tu, Siwei Ma, and Guangtao Zhai.
\newblock {Blind quality assessment for in-the-wild images via hierarchical
  feature fusion and iterative mixed database training}.
\newblock {\em IEEE Journal of Selected Topics in Signal Processing}, pages
  1--15, 2023.

\bibitem{sun2020mc360iqa}
Wei Sun, Xiongkuo Min, Guangtao Zhai, Ke Gu, Huiyu Duan, and Siwei Ma.
\newblock {MC360IQA: A multi-channel CNN for blind 360-degree image quality
  assessment}.
\newblock {\em IEEE Journal of Selected Topics in Signal Processing},
  14(1):64--77, 2020.

\bibitem{teixeira2016new}
Raoni~FS Teixeira and Neucimar~J Leite.
\newblock {A new framework for quality assessment of high-resolution
  fingerprint images}.
\newblock {\em IEEE Transactions on Pattern Analysis and Machine Intelligence},
  39(10):1905--1917, 2016.

\bibitem{tian2020light}
Yu Tian, Huanqiang Zeng, Junhui Hou, Jing Chen, Jianqing Zhu, and Kai-Kuang Ma.
\newblock {A light field image quality assessment model based on symmetry and
  depth features}.
\newblock {\em IEEE Transactions on Circuits and Systems for Video Technology},
  31(5):2046--2050, 2021.

\bibitem{upenik2016testbed}
Evgeniy Upenik, Martin {\v{R}}e{\v{r}}{\'a}bek, and Touradj Ebrahimi.
\newblock {Testbed for subjective evaluation of omnidirectional visual
  content}.
\newblock In {\em IEEE Picture Coding Symposium (PCS)}, pages 1--5, 2016.

\bibitem{vinyals2016show}
Oriol Vinyals, Alexander Toshev, Samy Bengio, and Dumitru Erhan.
\newblock {Show and tell: Lessons learned from the 2015 mscoco image captioning
  challenge}.
\newblock {\em IEEE Transactions on Pattern Analysis and Machine Intelligence},
  39(4):652--663, 2016.

\bibitem{waltl2010improving}
Markus Waltl, Christian Timmerer, and Hermann Hellwagner.
\newblock {Improving the quality of multimedia experience through sensory
  effects}.
\newblock In {\em IEEE International Workshop on Quality of Multimedia
  Experience (QoMEX)}, pages 124--129, 2010.

\bibitem{wang2022low}
Miaohui Wang, Yijing Huang, Jian Xiong, and Wuyuan Xie.
\newblock {Low-light Images In-the-wild: A Novel Visibility Perception-guided
  Blind Quality Indicator}.
\newblock {\em IEEE Transactions on Industrial Informatics}, 1(1):1--11, 2022.

\bibitem{wang2021blind}
Miaohui Wang, Yijing Huang, and Jialin Zhang.
\newblock {Blind Quality Assessment of Night-Time Images Via Weak Illumination
  Analysis}.
\newblock In {\em IEEE International Conference on Multimedia and Expo (ICME)},
  pages 1--6, 2021.

\bibitem{wang2022super}
Miaohui Wang, Zhuowei Xu, Yuanhao Gong, and Wuyuan Xie.
\newblock {S-CCR: Super-Complete Comparative Representation for Low-Light Image
  Quality Inference In-the-wild}.
\newblock In {\em ACM International Conference on Multimedia (MM)}, pages
  5219--5227, 2022.

\bibitem{wang2023blind}
Miaohui Wang, Zhuowei Xu, Mai Xu, and Weisi Lin.
\newblock {Blind Multimodal Quality Assessment of Low-light Images}.
\newblock {\em arXiv preprint arXiv:2303.10369}, pages 1--15, 2023.

\bibitem{wang2016subjective}
Shiqi Wang, Ke Gu, Xiang Zhang, Weisi Lin, Li Zhang, Siwei Ma, and Wen Gao.
\newblock {Subjective and objective quality assessment of compressed screen
  content images}.
\newblock {\em IEEE Journal on Emerging and Selected Topics in Circuits and
  Systems}, 6(4):532--543, 2016.

\bibitem{wang2022combining}
Xuejin Wang, Feng Shao, Qiuping Jiang, Zhenqi Fu, Xiangchao Meng, Ke Gu, and
  Yo-Sung Ho.
\newblock {Combining Retargeting Quality and Depth Perception Measures for
  Quality Evaluation of Retargeted Stereopairs}.
\newblock {\em IEEE Transactions on Multimedia}, 24(1):2422--2434, 2022.

\bibitem{wang2021active}
Zhihua Wang and Kede Ma.
\newblock {Active fine-tuning from gMAD examples improves blind image quality
  assessment}.
\newblock {\em IEEE Transactions on Pattern Analysis and Machine Intelligence},
  1(1):1--14, 2021.

\bibitem{wei2022contrastive}
Xuekai Wei, Jing Li, Mingliang Zhou, and Xianmin Wang.
\newblock {Contrastive distortion-level learning-based no-reference
  image-quality assessment}.
\newblock {\em Wiley International Journal of Intelligent Systems},
  37(11):8730--8746, 2022.

\bibitem{wen2020visual}
Wenying Wen, Kangkang Wei, Yuming Fang, and Yushu Zhang.
\newblock {Visual quality assessment for perceptually encrypted light field
  images}.
\newblock {\em IEEE Transactions on Circuits and Systems for Video Technology},
  31(7):2522--2534, 2020.

\bibitem{winkler2006perceived}
Stefan Winkler and Christof Faller.
\newblock {Perceived audiovisual quality of low-bitrate multimedia content}.
\newblock {\em IEEE transactions on multimedia}, 8(5):973--980, 2006.

\bibitem{wu2020end}
Jinjian Wu, Jupo Ma, Fuhu Liang, Weisheng Dong, Guangming Shi, and Weisi Lin.
\newblock {End-to-end blind image quality prediction with cascaded deep neural
  network}.
\newblock {\em IEEE Transactions on Image Processing}, 29(1):7414--7426, 2020.

\bibitem{wu2020subjective}
Qingbo Wu, Lei Wang, King~Ngi Ngan, Hongliang Li, Fanman Meng, and Linfeng Xu.
\newblock {Subjective and objective de-raining quality assessment towards
  authentic rain image}.
\newblock {\em IEEE Transactions on Circuits and Systems for Video Technology},
  30(11):3883--3897, 2020.

\bibitem{xiang2020blind}
Tao Xiang, Ying Yang, and Shangwei Guo.
\newblock {Blind night-time image quality assessment: Subjective and objective
  approaches}.
\newblock {\em IEEE Transactions on Multimedia}, 22(5):1259--1272, 2019.

\bibitem{xie2023pmbqa}
Wuyuan Xie, Kaimin Wang, Yakun Ju, and Miaohui Wang.
\newblock pmbqa: Projection-based blind point cloud quality assessment via
  multimodal learning.
\newblock In {\em ACM International Conference on Multimedia (ACM MM)}, pages
  3250--3258, 2023.

\bibitem{xu2016blind}
Jingtao Xu, Peng Ye, Qiaohong Li, Haiqing Du, Yong Liu, and David Doermann.
\newblock {Blind image quality assessment based on high order statistics
  aggregation}.
\newblock {\em IEEE Transactions on Image Processing}, 25(9):4444--4457, 2016.

\bibitem{xu2022viewport}
Mai Xu, Lai Jiang, Chen Li, Zulin Wang, and Xiaoming Tao.
\newblock {Viewport-based CNN: A multi-task approach for assessing 360 video
  quality}.
\newblock {\em IEEE Transactions on Pattern Analysis and Machine Intelligence},
  44(4):2198--2215, 2022.

\bibitem{xu2017no}
Shaoping Xu, Shunliang Jiang, and Weidong Min.
\newblock {No-reference/blind image quality assessment: A survey}.
\newblock {\em Taylor \& Francis IETE Technical Review}, 34(3):223--245, 2017.

\bibitem{xue2014blind}
Wufeng Xue, Xuanqin Mou, Lei Zhang, Alan~C Bovik, and Xiangchu Feng.
\newblock {Blind image quality assessment using joint statistics of gradient
  magnitude and Laplacian features}.
\newblock {\em IEEE Transactions on Image Processing}, 23(11):4850--4862, 2014.

\bibitem{yan2019naturalness}
Bo Yan, Bahetiyaer Bare, and Weimin Tan.
\newblock {Naturalness-aware deep no-reference image quality assessment}.
\newblock {\em IEEE Transactions on Multimedia}, 21(10):2603--2615, 2019.

\bibitem{yang2015perceptual}
Huan Yang, Yuming Fang, and Weisi Lin.
\newblock {Perceptual quality assessment of screen content images}.
\newblock {\em IEEE Transactions on Image Processing}, 24(11):4408--4421, 2015.

\bibitem{yang2022fine}
Wen Yang, Jinjian Wu, Shiwei Tian, Leida Li, Weisheng Dong, and Guangming Shi.
\newblock {Fine-Grained Image Quality Caption With Hierarchical Semantics
  Degradation}.
\newblock {\em IEEE Transactions on Image Processing}, 31(1):3578--3590, 2022.

\bibitem{yang2019survey}
Xiaohan Yang, Fan Li, and Hantao Liu.
\newblock {A survey of DNN methods for blind image quality assessment}.
\newblock {\em IEEE Access}, 7:123788--123806, 2019.

\bibitem{yeganeh2012objective}
Hojatollah Yeganeh and Zhou Wang.
\newblock {Objective quality assessment of tone-mapped images}.
\newblock {\em IEEE Transactions on Image processing}, 22(2):657--667, 2012.

\bibitem{ying2022telepresence}
Zhenqiang Ying, Deepti Ghadiyaram, and Alan Bovik.
\newblock {Telepresence video quality assessment}.
\newblock In {\em Springer European Conference on Computer Vision (ECCV)},
  pages 327--347, 2022.

\bibitem{ying2020patches}
Zhenqiang Ying, Haoran Niu, Praful Gupta, Dhruv Mahajan, Deepti Ghadiyaram, and
  Alan~C. Bovik.
\newblock {From patches to pictures (PaQ-2-PiQ): Mapping the perceptual space
  of picture quality}.
\newblock In {\em IEEE Conference on Computer Vision and Pattern Recognition
  (CVPR)}, pages 3575--3585, 2020.

\bibitem{yu2023hybrid}
Shaode Yu, Jiayi Wang, Jiacheng Gu, Mingxue Jin, Yunling Ma, Lijuan Yang, and
  Jianguang Li.
\newblock {A hybrid indicator for realistic blurred image quality assessment}.
\newblock {\em Elsevier Journal of Visual Communication and Image
  Representation}, 94:103848, 2023.

\bibitem{yue2019no}
Guanghui Yue, Chunping Hou, Ke Gu, Tianwei Zhou, and Hantao Liu.
\newblock {No-reference quality evaluator of transparently encrypted images}.
\newblock {\em IEEE Transactions on Multimedia}, 21(9):2184--2194, 2019.

\bibitem{zerman2017extensive}
Emin Zerman, Giuseppe Valenzise, and Frederic Dufaux.
\newblock {An extensive performance evaluation of full-reference HDR image
  quality metrics}.
\newblock {\em Springer Quality and User Experience}, 2(1):1--16, 2017.

\bibitem{zhai2020perceptual}
Guangtao Zhai and Xiongkuo Min.
\newblock {Perceptual image quality assessment: A survey}.
\newblock {\em Springer Science China Information Sciences}, 63(11):1--52,
  2020.

\bibitem{zhai2012psychovisual}
Guangtao Zhai, Xiaolin Wu, Xiaokang Yang, Weisi Lin, and Wenjun Zhang.
\newblock {A psychovisual quality metric in free-energy principle}.
\newblock {\em IEEE Transactions on Image Processing}, 21(1):41--52, 2012.

\bibitem{zhang2015feature}
Lin Zhang, Lei Zhang, and Alan~C Bovik.
\newblock {A feature-enriched completely blind image quality evaluator}.
\newblock {\em IEEE Transactions on Image Processing}, 24(8):2579--2591, 2015.

\bibitem{zhang2015application}
Wei Zhang, Ali Borji, Zhou Wang, Patrick Le~Callet, and Hantao Liu.
\newblock {The application of visual saliency models in objective image quality
  assessment: A statistical evaluation}.
\newblock {\em IEEE Transactions on Neural Networks and Learning Systems},
  27(6):1266--1278, 2015.

\bibitem{zhang2022continual}
Weixia Zhang, Dingquan Li, Chao Ma, Guangtao Zhai, Xiaokang Yang, and Kede Ma.
\newblock {Continual learning for blind image quality assessment}.
\newblock {\em IEEE Transactions on Pattern Analysis and Machine Intelligence},
  1(1):1--15, 2022.

\bibitem{zhang2020blind}
Weixia Zhang, Kede Ma, Jia Yan, Dexiang Deng, and Zhou Wang.
\newblock {Blind image quality assessment using a deep bilinear convolutional
  neural network}.
\newblock {\em IEEE Transactions on Circuits and Systems for Video Technology},
  30(1):36--47, 2020.

\bibitem{zhang2021uncertainty}
Weixia Zhang, Kede Ma, Guangtao Zhai, and Xiaokang Yang.
\newblock {Uncertainty-aware blind image quality assessment in the laboratory
  and wild}.
\newblock {\em IEEE Transactions on Image Processing}, 30(1):3474--3486, 2021.

\bibitem{zhao2023quality}
Kai Zhao, Kun Yuan, Ming Sun, Mading Li, and Xing Wen.
\newblock Quality-aware pre-trained models for blind image quality assessment.
\newblock In {\em IEEE Conference on Computer Vision and Pattern Recognition
  (CVPR)}, pages 22302--22313, 2023.

\bibitem{zheng2019no}
Linru Zheng, Liquan Shen, Jianan Chen, Ping An, and Jun Luo.
\newblock {No-reference quality assessment for screen content images based on
  hybrid region features fusion}.
\newblock {\em IEEE Transactions on Multimedia}, 21(8):2057--2070, 2019.

\bibitem{zheng2022uif}
Yannan Zheng, Weiling Chen, Rongfu Lin, Tiesong Zhao, and Patrick Le~Callet.
\newblock Uif: An objective quality assessment for underwater image
  enhancement.
\newblock {\em IEEE Transactions on Image Processing}, 1:5456--5468, 2022.

\bibitem{zhou2019visual}
Fei Zhou, Rongguo Yao, Bozhi Liu, and Guoping Qiu.
\newblock {Visual quality assessment for super-resolved images: Database and
  method}.
\newblock {\em IEEE Transactions on Image Processing}, 28(7):3528--3541, 2019.

\bibitem{zhou2022no}
Wei Zhou, Jiahua Xu, Qiuping Jiang, and Zhibo Chen.
\newblock {No-reference quality assessment for 360-degree images by analysis of
  multifrequency information and local-global naturalness}.
\newblock {\em IEEE Transactions on Circuits and Systems for Video Technology},
  32(4):1778--1791, 2022.

\bibitem{zhu2020metaiqa}
Hancheng Zhu, Leida Li, Jinjian Wu, Weisheng Dong, and Guangming Shi.
\newblock {MetaIQA: Deep meta-learning for no-reference image quality
  assessment}.
\newblock In {\em IEEE Conference on Computer Vision and Pattern Recognition
  (CVPR)}, pages 14143--14152, 2020.

\end{thebibliography}
}

\end{document}